\documentclass[journal]{IEEEtran}  

\IEEEoverridecommandlockouts                              

\usepackage[pdftex]{graphicx}
\DeclareGraphicsExtensions{.pdf,.jpeg,.png}
\usepackage{amsmath} 
\usepackage{amssymb}  

\usepackage{color,soul}
\usepackage{tabularx,booktabs,multirow}
\usepackage{array}
\usepackage{float}
\usepackage{makecell}
\usepackage{tablefootnote}
\usepackage{multicol}
\usepackage{booktabs}
\usepackage{threeparttable}
\usepackage{url}
\usepackage[font=small,skip=10pt]{caption}
\usepackage[dvipsnames]{xcolor}
\usepackage{amsmath}
\usepackage{xcolor}
\usepackage{algorithmic}
\usepackage{dirtytalk}
\usepackage{hyperref}
\DeclareMathOperator*{\argminA}{arg\,min}
\usepackage{framed}
\usepackage{enumitem}
\hypersetup{colorlinks,allcolors=red}
\usepackage[ruled,linesnumbered]{algorithm2e}
\newcommand{\R}{\mathbb{R}}
\newcolumntype{P}[1]{>{\centering\arraybackslash}p{#1}}
\SetKwInOut{Input}{input}
\SetKwInOut{Output}{output}

\title{
Formulating Intuitive Stack-of-Tasks using Visuo-Tactile Perception for Collaborative Human-Robot Fine Manipulation
}

\author{Sunny Katyara$^{1,2}$, Nikhil Deshpande$^{1}$,~\IEEEmembership{Member,~IEEE}, Fanny Ficuciello$^{2}$,~\IEEEmembership{Senior Member,~IEEE}, Tao Teng$^{1}$, Bruno Siciliano$^{2}$,~\IEEEmembership{Fellow,~IEEE}, Darwin G. Caldwell$^{1}$,~\IEEEmembership{Senior Member,~IEEE}, Fei Chen$^{3}$,~\IEEEmembership{Senior Member,~IEEE}
\thanks{This research is supported by the projects – “LEARN-REAL” funded by EU H2020 ERA-Net Chist-Era; “HARMONY” funded by EU H2020 R\&I under agreement No. 101017008; and by the Italian Workers’ Compensation Authority (INAIL) under the “Sistemi Cibernetici Collaborativi” agreement (INAIL)  \textit{(Corresponding author: Darwin G. Caldwell).} }
\thanks{$^{1}$ APRIL Lab, Department of Advanced Robotics, Istituto Italiano di Tecnologia, Genova, Italy (e-mail: {\tt\small name.surname@iit.it}).}
\thanks{$^{2}$ Department of Information Technology and Electrical Engineering and PRISMA Lab, University of Naples Federico II, Naples, Italy ({e-mail: \tt\small name.surname@unina.it}).}
\thanks{$^{3}$ Department of Mechanical and Automation Engineering, T-Stone Robotics Institute, The Chinese University of Hong Kong, Hong Kong ({e-mail: \tt\small f.chen@ieee.org}).}
}

\begin{document}

\maketitle
\IEEEdisplaynontitleabstractindextext
\IEEEpeerreviewmaketitle

\begin{abstract}

Enabling robots to work in close proximity to humans necessitates a control framework that does not only incorporate multi-sensory information for autonomous and coordinated interactions but also has perceptive task planning to ensure an adaptable and flexible collaborative behaviour. In this research, an intuitive stack-of-tasks (iSoT) formulation is proposed, that defines the robot's actions by considering the human-arm postures and the task progression. The framework is augmented with visuo-tactile information to effectively perceive the collaborative environment and intuitively switch between the planned sub-tasks. The visual feedback from depth cameras monitors and estimates the objects' poses and human-arm postures, while the tactile data provides the exploration skills to detect and maintain the desired contacts to avoid object slippage. To evaluate the performance, effectiveness and usability of the proposed framework, assembly and disassembly tasks, performed by the human-human and human-robot partners, are considered and analyzed using distinct evaluation metrics i.e, approach adaptation, grasp correction, task coordination latency, cumulative posture deviation, and task repeatability.

\end{abstract}


\section{INTRODUCTION}

\begin{figure}[t]
      \centering
      \includegraphics[width=8.5cm]{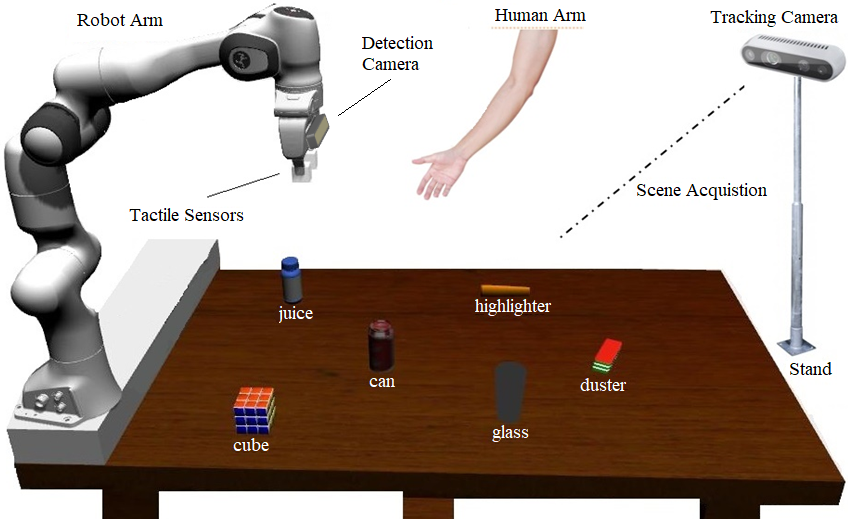}
      \caption{The proposed setup on using visuo-tactile feedback to achieve an intuitive human--robot collaboration. The visual information detects the human and objects in the environment and estimates their postures and poses respectively. The tactile sensing updates the force profile of the gripper during task execution for flexible interactions.}
      \label{one}
      \vspace{-15pt}
   \end{figure}

\IEEEPARstart{H}{uman Robot Collaboration} (HRC) --- a domain that combines the cognitive and physical capabilities of humans with the strength and precision of robots to achieve a common goal --- has emerged from well constrained industrial settings to boundary-less environments, where it can provide shared assistance to the humans \cite{c1}\cite{c2}. Humans, who are excellent at performing fine operations such as; assembling/disassembling, aligning, untying etc, when teamed up with the collaborative robots, that have been designed to repeat the given tasks with higher flexibility, accuracy and precision, essentially improve the quality and efficiency of the industrial process \cite{c3}. However, HRC in industrial settings still faces numerous challenges especially with respect to the absolute safety of co-workers, the robustness of interaction mechanisms, the ease of use, and user satisfaction \cite{c4}. Addressing these issues requires a detailed understanding of the dynamic constraints arising during the collaborative tasks such as; the nature of task, the intention of interaction, the human postures, the pose of objects and many others. In particular, the robot first needs to recognize the presence of human i.e, its postures, to assume a desired configuration, and then identify candidate objects in the environment to interact and adapt to them in a reactive way. In this domain, the exteroceptive sensor-based techniques, involving visual servoing and haptic exploration, are promising strategies for achieving an intuitive HRC (iHRC), as depicted in Fig. \ref{one} as compared to planning based approaches using proprioceptive information, where the knowledge of agents and environment are encoded manually beforehand, which are highly uncertain practically \cite{c5}\cite{c6}\cite{c7}.  

   \begin{figure*}[t]
      \centering
      \includegraphics[width=17cm]{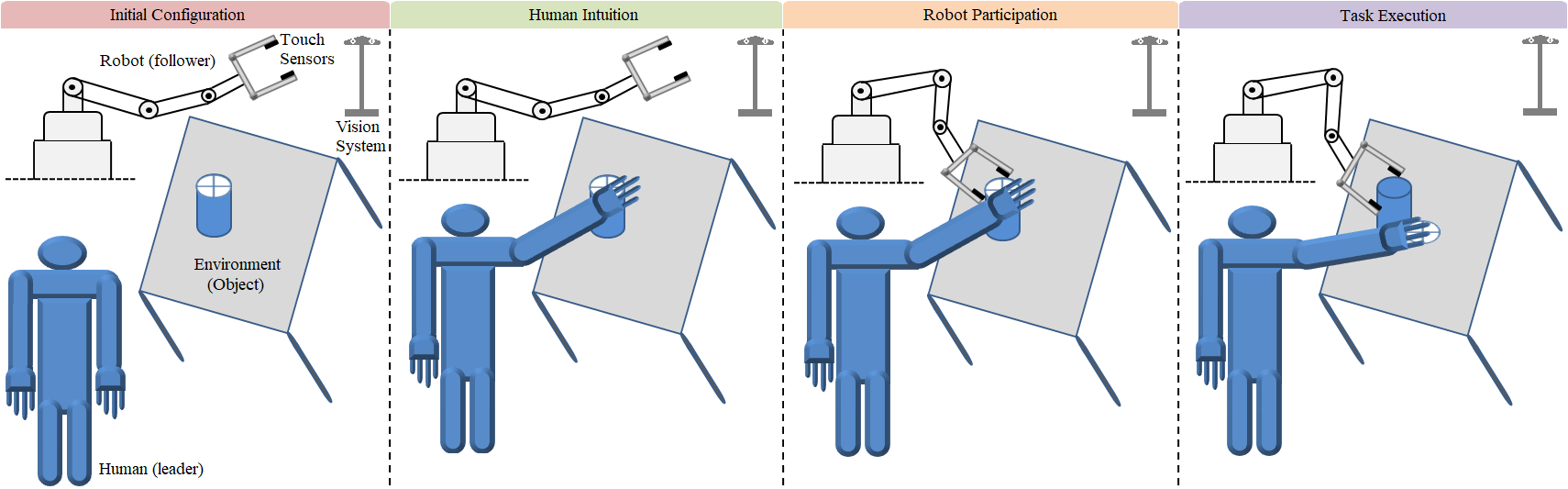}
      \caption{Sketch of intuitive human-robot collaboration, Initial Configuration -- represents the homing position of all the actors i.e., human (leader), robot (follower) and environment (object) in a collaborative setup, Human Intuition -- is a condition where human intents to open the jar with its hand grasping the lid, Robot Participation -- is a condition where robot follows the human intention using a feedback from the vision system and assists by gripping the body of jar, and Task Execution -- is a condition where human applies outward force to remove the lid while the robot is holding the jar using its gripper coupled with the touch sensors to avoid slippage.}
      \label{two}
      \vspace{-15pt}
   \end{figure*}

The iHRC refers to a condition where a robot partner interprets the gestures or intentions of a user to understand his/her interactions and subsequently collaborate more naturally from the human viewpoint \cite{c8}\cite{c9}. The iHRC is a blend of contact-less and physical collaborations to not only ensure a safe coexistence but also to achieve coordinated motions and actions. Where, the visual servoing assists to coordinate the relative motions of human-robot partners and the haptic exploration helps to estimate their interaction forces to react accordingly, as graphically illustrated in Fig. \ref{two}. In this scenario, a human subject aims to perform a task that invariably requires the actions and co-ordinations of the dual arms (that can be either the human--human or human--robot case). In this iHRC task, the tactile (haptic) information ensures a flexible control of interaction forces between the human and robot arms, and the common object (i.e, jar) in a collaborative setup. However, only the sense of touch is not enough to perform this task and thus, the visual feedback is introduced into the system to complement the tactile data for coordinated motion control based on the human-arm's postures, objects' poses and task progression in the collaborative environment \cite{c10}\cite{c11}. 

Fusing visuo-tactile perception into the control architecture to achieve a coordinated motion control entails to implement one of three control strategies i.e., traded, hybrid and shared topologies \cite{c12}. In the traded scheme, based on the magnitude of task error threshold, there is a switching between the visual servoing and tactile exploration. The hybrid approach, on the other hand, combines both the sensing modalities but decouples them in an orthogonal space for an independent use thereby considering the description of respective task frames. Finally, the shared strategy utilizes both the forms of sensory information simultaneously in the same sub-space.

The reminder of this paper is organized as follows. Section II reviews the related research work carried out in the field of HRC exploiting either visual or/and haptic information for different industrial and commercial scenarios and outlines the objectives of this research. Section III presents the formulation of the proposed intuitive stack-of-tasks framework thereby exploiting the visuo-tactile feedback. Section IV discusses and analyzes the experiments performed using the proposed framework for assembly and disassembly tasks. Section V gives a comparative analysis between the similar tasks performed by the human-human and human-robot partners. Section VI draws conclusions on the key outcomes of this research and proposes possible extensions for the future work.
   
\section{Related Works and Research Objectives}

To exploit visual and haptic information to produce flexible and reactive coordination between a robot and human, a hybrid control scheme is proposed in \cite{c13} for transportation tasks. The method employs a vision system that monitors the status of the task using a minimum jerk model, while the haptics ensures a compliant and an adaptive interaction between the two agents. But the test scenario is very regulated, operating in a fully structured environment, with markers being used to track the human motions. A marker-less traded control framework for a coordinated assembly task is presented in \cite{c14}. This technique is robust against task variations and changing human postures, but it does not exclusively utilize the haptics for force modulation during task execution that eventually result into poor task performance and, also, it assumes that the pose of a given tool is known. A hybrid control scheme, to determine the coordinated movements of human-robot and also track the status of task using visual and haptic information under an unstructured environment, is discussed in \cite{c15}. In this method, the robot actions are defined using the notion of stack-of-tasks, but without considering the active human postures. Following the same principle, a unified shared control formulation is described in \cite{c16} for an industrial co-manipulation scenario. Apart from using visual and haptic feedback, it incorporates adaptive gains and homotopy to generate a smooth transition between the human and robot actions, but it does not take into account the dynamic variations in the agents’ behaviour and the environment, to switch intuitively. To address this, a traded architecture on fusing force/torque and depth information for run-time intention monitoring, detection and tool regulation under controlled industrial assembly settings is proposed in  \cite{c17}. The presented approach is a multi-objective model with hard constraints on safe coexistence and flexible collaboration but is computationaly ill conditioned and is not a cost effective solution for regular commercial and industrial tasks. A hybrid framework is proposed in \cite{c18} for a commercial handover task under an unstructured environment considering the robot--robot and human--robot topologies. The proposed formulation fuses a dynamic reactive controller with the event driven decision-making to enable the robot to adapt to the humans' behaviour and their transient responses. The approach, while being effective against dynamic variations, is not extensively scalable and task agnostic. To monitor the progress of unknown tasks and plan the desired actions accordingly, a traded strategy is discussed in \cite{c19}. The method uses a wearable haptic device to inform the human coworker about the robot's actions prior to their execution. The robot then plans a collision free path using the aligned depth-map information. The idea is simple and effective for industrial scenarios but it eventually increases the cognitive burden and also has slow task performance due to a significant communication latency. Considering the limitations of above techniques, this paper proposes an intuitive control framework that is not only robust and optimal but also exploits human-arm postures and task progression to formulate the hierarchical actions of the robot partner using visuo-tactile feedback for improving the planning, performance and execution of the collaborative fine manipulation tasks.

Overall, the key objectives of this paper are; \textbf{(i)} to formulate an intuitive stack-of-tasks (iSoT) framework by taking into account the human-arm postures and the task progression thereby using visual and tactile perceptions in a traded fashion, \textbf{(ii)} to evaluate the performance of the proposed framework on assembly and disassembly tasks i.e., screwing a bolt into a nut and detaching a marker from a cap, \textbf{(iii)} to analyze the effectiveness and usability of the proposed framework (that is designed for HRC) using distinct evaluation metrics i.e., approach adaptation, grasp correction, task coordination latency, cumulative posture deviation, and task repeatability, against the human--human cooperation (HHC).   

\section{Proposed Framework}

   \begin{figure}[t]
      \centering
      \includegraphics[height=4.25cm, width=8.5cm]{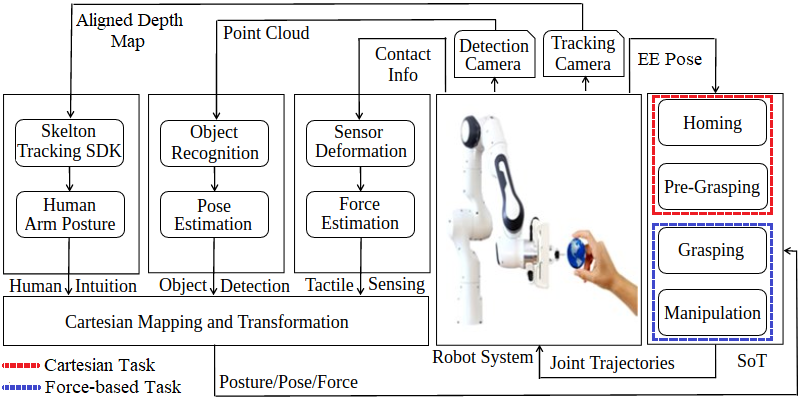}
      \caption{The proposed intuitive stack-of-tasks framework using tactile sensing, object detection and human intuition in traded fashion.}
      \label{three}
      \vspace{-15pt}
   \end{figure}

\begin{figure}[t]
      \centering
      \includegraphics[width=8.5cm]{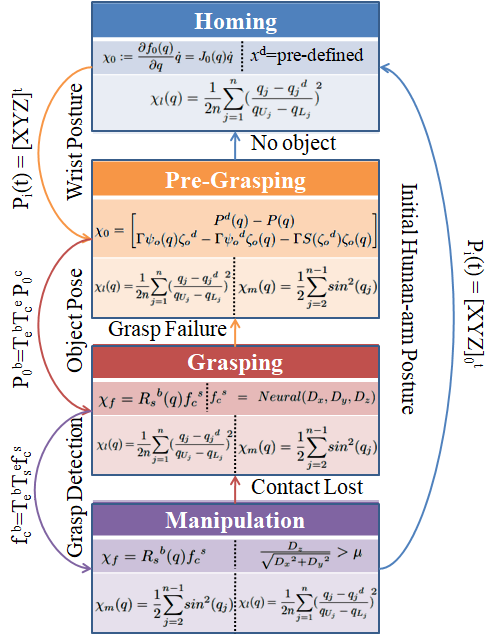}
      \caption{The intuitive switching strategy for transition between the sub-tasks within the formulated stack-of-tasks.}
      \label{three_A}
      \vspace{-15pt}
   \end{figure}

The block diagram of the proposed framework is shown in Fig. \ref{three}, where the stack-of-tasks (SoT) is defined for the Cartesian (i.e, homing and pre-grasping) and the force (i.e, grasping and manipulation) based tasks as a series of Quadratic Programming (QP) problems. Each task has primary and secondary descriptions based on the nature and constraints of the robot's actions. For the tasks being defined, the switching between them occurs intuitively based on the human-arm postures and task progression using the visual (aligned depth map and point cloud) and the tactile (contact info) information, as shown in Fig. \ref{three_A}. The depth data monitors the human-arm postures (using a tracking camera) as well as it detects the candidate objects in the scene (using a detection camera). The tactile sensing helps in modulating the force profile of the gripper during grasping and manipulation phases. This ensures to avoid the slippage of grasped object. The flow chart (switching strategy) in Fig. \ref{three_A} states that the robot for each HRC task starts in the homing position and switches to a pre-grasping configuration when prompted by the posture of active human arm (i.e., the one involved in human-robot interaction) using the tracking camera, but it withdraws if no candidate object is found in the collaborative environment. From pre-grasping, the robot enters the grasping phase based on the pose information of the segmented candidate object using the detection camera, but it re-assumes the pre-grasping configuration if there is a grasp failure. The transition to manipulation state occurs when the magnitude of the tactile feedback fulfills the friction cone criteria (for grasp stability), but if this does not occur, then the robot returns to the grasping phase for re-execution (i.e, contact recovery). Finally, the robot returns to its homing configuration using a feedback from the tracking camera that ensures that the human-arm has retained its initial posture (i.e, a leader--follower paradigm).

\subsection{Stack-of-Tasks Formulation}

  \begin{figure*}[t]
      \centering
      \includegraphics[width=17cm]{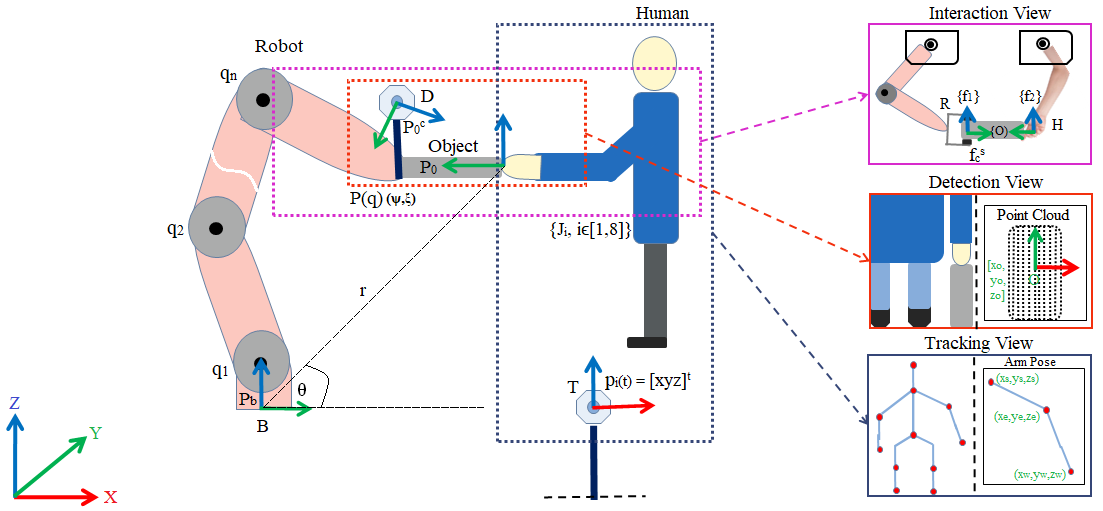}
      \caption{Characterization of collaborative scenario considering the respective frame transformations of human (leader), robot (follower) and environment (object) into the robot base frame. With the robot and human coordinating their actions onto a common object, the three key perspectives are considered i.e., the tracking view T (monitors the human-arm posture), the detection view D (localizes the pose of object), the interaction view I (considers the possible forces exerted by the human and robot onto a common object).}
      \label{four}
      \vspace{-15pt}
  \end{figure*}

Let $q=[q_1,q_2,....,q_n]\in\R^{n}$ be the vector of robot arm joints describing its possible configurations. Then, the primary Cartesian task, defined by $\chi_0:=f_0(q)\in\R^{\pi_0}$, can be expressed in terms of the differential kinematics as 

\begin{equation}
    \dot{\chi_0}:=\frac{\partial f_0(q)}{\partial q}\Dot{q}=J_0(q)\Dot{q}
\label{eqt_1}
\end{equation}

where $n$ and $\pi_0$ are the degrees of freedom of the robot arm and the dimensions of the primary task respectively, and $J_0(q)\in\R^{{\pi_0}\times{n}}$ is the primary task Jacobian. To ensure asymptotic convergence of task $\dot{\chi_0}$ to its desired value $\dot{\chi_0}^{d}$, an error function $e_0=\dot{\chi_0}^{d}-\dot{\chi_0}$ is defined. Hence, the vector of joint velocities is determined by inverting (\ref{eqt_1})

\begin{equation}
    \Dot{q}^{d}=J_0^{\dagger}(q){\Omega_0}{e_0}
\label{eqt_2}
\end{equation}

where $\Omega_0\in{\R^{\pi_0\times\pi_0}}$ is a positive-definite gain matrix (i.e, chosen empirically for task convergence) and $J_0^{\dagger}(q)$ is the pseudo inverse of the primary task Jacobian. 

Since the robot arm is kinematically redundant i.e., $n>\pi_0$, the null space of the primary task Jacobian $J_0(q)$ is non-trivial and thus can be exploited to define the secondary tasks $\chi_1:=f_1(q)\in\R^{\pi_1}$ provided that $n-\pi_0\geq{\pi_1}$. In such a case, the vector of joint velocities is thus given by

\begin{equation}
    \Dot{q}^{d}=J_0^{\dagger}(q){\Omega_0}{e_0}+{N_0(q)}J_1^{\dagger}(q){\Omega_1}{e_1}
\label{eqt_3}
\end{equation}

where $N_0(q)={I_n}-J_0^{\dagger}(q)J_0(q)$ is a null space projector of the primary task Jacobian $J_0(q)$, $I_n\in\R^{n\times{n}}$ is the identity matrix, $J_1(q)\in{\R^{\pi_1\times{n}}}$ is the secondary task Jacobian. The formulation can be generalized to K-prioritized tasks thereby considering the task description and kinematic constraints of the robot arm configurations by using \cite{c20}. 

\begin{equation}
    \Dot{q}^{d}=J_0^{\dagger}(q){\Omega_0}{e_0}+{\sum_{j=1}^{K}}{N_{aug}(q)}J_j^{\dagger}(q){\Omega_j}{e_j}
\label{eqt_4}
\end{equation}

The matrix $N_{aug}(q)$ in (\ref{eqt_4}) is a null space projector of the augmented task Jacobian $J_{aug}$ with the corresponding task error function $e_{aug}$, defined in (\ref{eqt_5}), where $\alpha$ is a relative weight matrix, defining the priority level of the tasks. The greater the value of $\alpha$, the higher the priority of the respective task and vice-versa. Moreover, the inclusion of $\alpha$ also ensures that the matrix $J_{aug}$ does not loses it rank and thus is less prone to algorithmic singularity during the task execution.    

\begin{equation}
  \begin{array}{l}
    J_{aug}(q) = [{\alpha_0}J_0^{T}(q),.......,{\alpha_K}J_K^{T}(q)]^{T}
    \\
    e_{aug} = [{\alpha_0}e_0^{T},.......,{\alpha_K}e_K^{T}]^{T}
  \end{array}
\label{eqt_5}
\end{equation}

The general description of the primary and secondary tasks in (\ref{eqt_4}) represents their positioning and performance characteristics respectively. The positioning tasks are defined to maintain the position and orientation of the end-effector in a desired configuration according to a co-manipulation scenario whereas the performance tasks are designed to ensure an optimal behaviour of the robot arm. 

Let $J(q)\in\R^{m\times{n}}$ be the Jacobian of the robot arm that consists of position $J_p(q)\in\R^{3\times{n}}$ and orientation $J_o(q)\in\R^{(m-3)\times{n}}$ parts that are associated with the Cartesian pose information of the end-effector. Let $P(q)\in{\R^3}$ with $\{\psi_o(q),\zeta_o(q)\}$ and $P^{d}(q)\in\R^{3}$ with $\{{\psi_o}^{d},{\zeta_o}^{d}\}$ be the current and desired positions with unit quaternions of the end-effector in the robot base frame respectively, as illustrated in Fig. \ref{four}. Then, the primary Cartesian task can be defined as \cite{c21}

\begin{equation}
    \chi_0 = \begin{bmatrix}
 P^{d}(q)-P(q) \\ \Gamma{\psi_o}(q){\zeta_o}^{d}-\Gamma{\psi_o}^{d}{\zeta_o}(q)-\Gamma{S({\zeta_o}^{d})}{\zeta_o}(q)
\end{bmatrix}
\label{eqt_6}
\end{equation}

where $S(.)$ is a skew-symmetric matrix. In order to ensure the convergence of the primary task in (\ref{eqt_6}), the desired task variables are ${\chi_0}^{d}=0$, $\Gamma=\begin{bmatrix}
 1 & 0 & 0 \\ 0 & 1 & 0 
\end{bmatrix}$, and the corresponding task Jacobian matrix is $J_0 = \begin{bmatrix} J_p(q) \\ \Gamma{J_o(q)} \end{bmatrix}$. The matrix $\Gamma$ gives a transformation between the geometric Jacobian of the robot arm $J(q)$ and the analytical Jacobian of the Cartesian task $J_0$. Such a description exclusively neglects the singularities associated with the choice of Euler angles and thus makes our formulation more robust and computationally efficient.     

For the force based primary tasks, the deformation data of the installed tactile sensors (i.e, mapped to corresponding force values) is exploited to achieve a compliant behaviour of the gripper, as shown in Fig. \ref{seven} (a). Let ${f_c}^{s}$ be the vector of contact forces in the sensor frame and ${R_s}^{b}(q)\in{SO(3)}$ be a rotation matrix defining a transformation from the sensors to the robot base frames, as shown in Fig. \ref{four}. Hence, the task function and the corresponding task Jacobian are characterized by $\chi_f = {R_s}^{b}(q){f_c}^{s}$ and $J_f=J_{g}(q)$ respectively, with $J_{g}(q)\in{\R}^{3\times{2}}$ representing the Jacobian of the gripper. For the assembly and disassembly tasks, it is desirable to maintain a stiff positional behaviour of the gripper while a common object is being grasped and manipulated by the human subject \cite{c22}\cite{c23}. Therefore, in case of such static equilibrium, there exist a kineto-static duality between the generalized forces and joints velocities and the relationship is defined as $\dot{q}=J^{T}f_e$ (i.e, with ideal dynamics of the robot arm). Where, $f_e$ is the applied vector of end-effector forces and is related to gripper forces ${f_c}^{s}$ by a rot-translation matrix ${T_s}^{e}(q)\in{SE(3)}$.    

For the secondary tasks, two of the most important performance metrics, i.e., the manipulability measure and the joint limit avoidance, are considered in this research. The manipulability measure quantifies the possible configurations of the robot arm to avoid kinematic singularities. Let $q$ be a vector of an arbitrary robot arm configuration, then the manipulability task function can be expressed as \cite{c24}.

\begin{equation}
    \chi_m(q) = \frac{1}{2}{\sum_{j=2}^{n-1}}{sin}^{2}(q_{j})
\label{eqt_7}
\end{equation}

The desired value ${\chi_m}^{d}$ for this task is chosen empirically from the valid configurations of the robot arm, by considering its reachable work-space (i.e., [$0.85$ $m$, $4.7124$ rad] is the kinematic reach of the robot arm used in this research) and as well as its kinematic structure (i.e, DH table for axes alignment), whereas the corresponding task Jacobian is defined as $J_m(q)=[\mathrm{cos(q_1)}\mathrm{sin(q_1)},.....,\mathrm{cos(q_n)}\mathrm{sin(q_n)}]$. However, the task function defined in (\ref{eqt_7}) does not correspond to full manipulability of the robot arm but is one of the potential candidates having same minima as of the original complex manipulability function i.e, $\sqrt{det[J(q)J^T(q)}$ \cite{c20}.

The joint limit avoidance task, which aims at keeping the robot arm joints away from their mechanical limits (i.e, bringing them close to central values), is defined as \cite{c20}.

\begin{equation}
    \chi_l(q) = \frac{1}{2n}{\sum_{j=1}^{n}}{(\frac{q_j-{q_j}^{d}}{q_{U_j}-q_{L_j}})^2}
\label{eqt_8}
\end{equation}

where ${q_U}\in{\R}^{n}$ and ${q_L}\in{\R}^{n}$ are the upper and lower mechanical joint limits respectively, ${q_j}^{d}=\frac{{q_U}_j+{q_L}_l}{2}$ is the mid-point of $j^{th}$ joint range, and the corresponding task Jacobian is defined as $J_l(q)=\frac{1}{n}[\frac{q_1-{q_1}^d}{q_{U_1}-q_{L_1}},....,\frac{q_n-{q_n}^d}{q_{U_n}-q_{L_n}}]$.  

For all the tasks (i.e, primary and secondary) being defined and formulated, in general each of them is governed by associated task Jacobian and error function and thus can be expressed by a couple i.e., $T_j=(J_j, e_j)$. For the large of number of such tasks to be executed at their best and with strict priorities by the redundant robot system, they can be stacked together according to their rank and execution level i.e., building the stack-of-tasks (SoT). Each task in SoT can be assigned with either hard or soft priorities. The tasks are said to have hard priorities if their execution is not deteriorated by the other tasks, while those with soft priorities may influence each other based on their degree of correlation, i.e., the relative weights $\alpha_j$. Hence, the primary tasks, either the Cartesian or force based, are assigned with hard priorities, whereas the secondary tasks, i.e., manipulability measure and joint limit avoidance are given soft priorities.

The optimality of the tasks defined in SoT can be achieved by considering each of them as a constrained Quadratic Programming (QP) problem to be solved in a cascaded manner recursively within the solution set $S_s$, as \cite{c25}  

\begin{equation}
\begin{aligned}
    \Dot{q_n} = {\argminA_{\dot{q\in{S_s}}}\frac{1}{2} ||{J_n}\dot{q}-\Omega_n{e_n}||^2}+\frac{1}{2}||\Lambda_n||^2
    \\
    s.t
    \\
    l_n\leq{\Lambda_n}{\Dot{q}}\leq{u_n}
\end{aligned}
\label{eqt_9}
\end{equation}

where $l_n$ and $u_n$ are the lower and upper bounds on the joint velocity constraint ${\Lambda_n}{\Dot{q}}$ and $\Lambda_n$ is a set of non-negative slack variables. This recursive formulation does not only reduce computational burden but also facilitate reordering and swapping of the defined tasks. Hence, the solution obtained is sent to the velocity based or integrated position based low-level robot controller using ${q}^{d}=q+\Dot{q}\Delta{t}$, where $\Delta{t}$ is the control loop rate and for our robot arm, this is $1$ $ms$.

\subsection{Gesture Monitoring and Posture Evaluation}

   \begin{figure}[t]
      \centering
      \includegraphics[width=8.5cm]{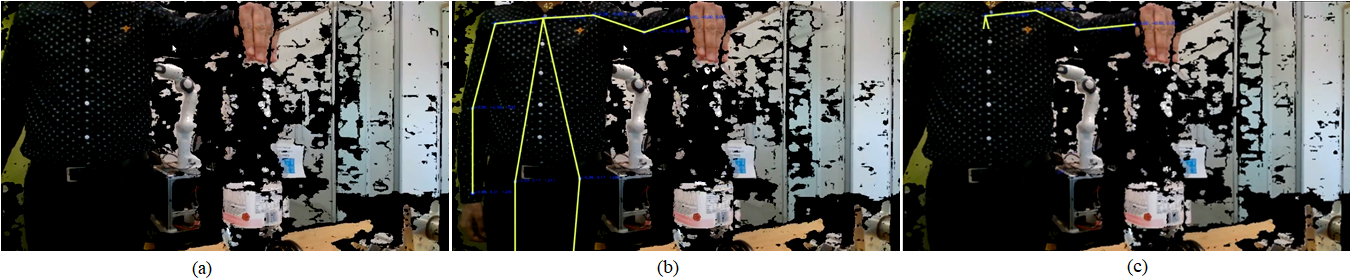}
      \caption{Estimating human-arm postures using the skeleton tracking algorithm, (a) represents the perspective view of a human performing certain actions, (b) illustrates the estimated poses of each of the 18 joints of a human in the camera frame, (c) shows the filtered tracking with only the poses of active human-arm joints (for precise human gesture monitoring during co-manipulation).\\}
      \label{eight}
      \vspace{-15pt}
   \end{figure}
   
To monitor the gesture of an active human-arm (leader) and evaluate its posture at run-time to command the robot arm (follower) either from or to its homing configuration in a co-manipulation scenario, the skeleton tracking algorithm is exploited \cite{c26}. The algorithm uses an aligned depth map, that is acquired using a depth camera (tracking camera), running at $5$ $Hz$, to detect a human subject in the environment, and tracks a set of his/her body joints in a 3D space. The algorithm (using deep CNN together with model based fitting) is a pre-trained model to estimate the poses $P_i(t)=[X_i Y_i Z_i]^{t}$ of 18 joints of the human body, where $t$ represents the reference frame of the detected joints. In particular, the joints of the active human arm (that are involved in co-manipulation) are localized using a customized band-pass filter for precise gesture monitoring and posture estimation. Mathematically, the pose of detected joints, mapped into camera Cartesian space (i.e, $P_{ci}=[X_{ci},Y_{ci},Z_{ci}]$) from the depth space, is given by \cite{c27} 

\begin{equation}
\begin{aligned}
X_{ci} = \frac{(I_{xi}-c_{xi})p_{di}}{f{s_{xi}}}, Y_{ci} = \frac{(I_{yi}-c_{yi})p_{di}}{f{s_{yi}}}, Z_{ci} = p_{di}
\end{aligned}
\label{eqt_40}
\end{equation}

where $I_x$ and $I_y$ are the pixel coordinates in the image plane, $c_x$ and $c_y$ represent the coordinates of center of focal axis, $p_d$ is the depth of point, f is the focal length of camera and, $s_x$ and $s_y$ are pixel dimensions. The corresponding value of $i^{th}$ joint pose into the robot base frame is determined using

\begin{equation}
\begin{aligned}
P_i = {R_c}^{b}P_{ci}+l
\end{aligned}
\label{eqt_41}
\end{equation}

where ${R_c}^{b}$ and $l$ represent rotation and translation between camera and robot base frames. Since, we are interested in the pose information of wrist $P_w$, that drives the error $e_0$ into (\ref{eqt_2}) and result into optimal configuration of robot arm $\dot{q}$ using (\ref{eqt_9}). 

An example of an aligned depth map of an environment, with a human subject and the tracked joints in the camera Cartesian space (both complete and filtered) using tracking algorithm, is shown in Fig. \ref{eight}. Where, the gray areas represent all the occluded points beyond the obstacles that form a region of uncertainty (i.e, based on the threshold imposed by $p_d$) for gesture monitoring and posture evaluation.

\subsection{Object Detection and Pose Estimation}

   \begin{figure}[t]
      \centering
      \includegraphics[width=8.5cm]{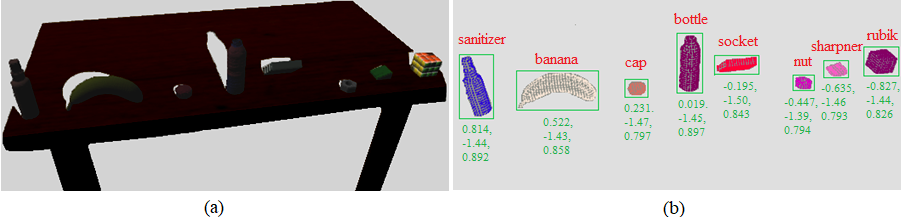}
      \caption{Object segmentation, recognition and pose estimation, (a) shows the raw point cloud of the scene, and (b) illustrates the estimated poses of recognized candidate objects, determined locally from the centroid of their respective processed point clouds.\\}
      \label{five}
      \vspace{-15pt}
   \end{figure}

To detect objects in the collaborative environment, the raw point cloud of the scene, shown in Fig. \ref{five}(a), is captured using the eye-in-hand (detection) camera, running at $25$ $Hz$. It is initially filtered and down sampled to remove adversarial data points and minimize the computational time respectively. Later, the standard RANSAC algorithm is applied to the processed point cloud to segment out the candidate objects from a scene using the Euclidean clustering criteria as \cite{c28}

   \begin{figure}[t]
      \centering
      \includegraphics[width=8.5cm]{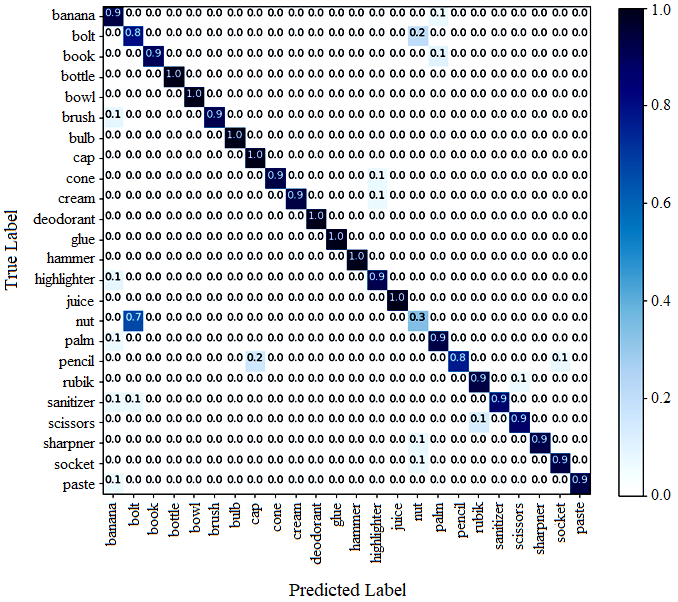}
      \caption{The normalized confusion matrix of SVM classifier for object recognition during model inference.\\}
      \label{confusion}
      \vspace{-15pt}
   \end{figure}

\begin{equation}
    O_c = {\int_{j=0}^{K}}N_{\xi}(x_j|MinPts):{\{y_j|d(x_j,y_j){\leq}{\xi}\}}dx
\label{eqt_10}
\end{equation}

where K represents the number of clusters generated, $\xi$ is the distance to neighbouring points $N_\xi$, and $MinPts$ is the minimum number of neighbouring points within the threshold of $\xi$. Two neighbouring points i.e., x and y can be grouped together if the distance between them is less than or equal to $\xi$ otherwise they are neglected. 

The candidate objects found in terms of clusters using (\ref{eqt_10}), are further classified using a trained SVM classifier and their poses are enumerated locally by computing the centroid (i.e., where the object's reference frame is defined) of their processed point cloud, as shown in Fig. \ref{five}(b). The SVM classifier has been trained on 24 distinct daily life objects with over 50 different orientations of each. This achieves a $84.3\%$ accuracy during the inference, as shown by the confusion matrix in Fig. \ref{confusion}. However, the enumerated poses of the detected objects are in the camera frame and must be transformed into the robot base frame using suitable transformations according to Fig. \ref{four}
   
\begin{equation}
    {P_o}^{b} = {T_e}^{b}{T_c}^{e}{P_o}^{c}
\label{eqt_11}
\end{equation}

where ${P_o}^{c}\in{\R^{4\times{1}}}$ and ${P_o}^{b}\in{\R^{4\times{1}}}$ are the object poses in the camera and the robot base frames respectively, while ${T_c}^{e}\in{SE(3)}$ and ${T_e}^{b}\in{SE(3)}$ are the homogeneous transformations between the in-hand camera and the end-effector, and the end-effector and the robot base frames respectively.

\subsection{Tactile Sensing and Force Mapping}

\begin{figure}[t]
      \centering
      \includegraphics[width=8.5cm]{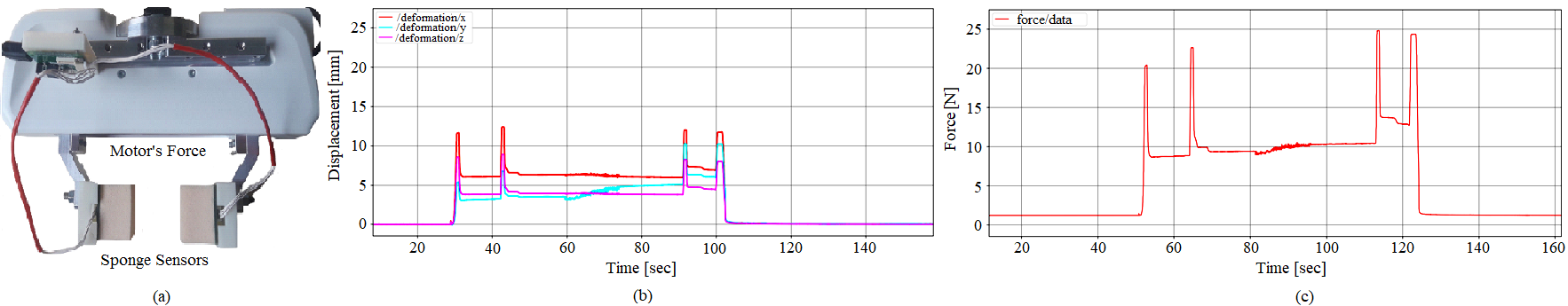}
      \caption{Tactile sensing and force profile of the gripper, (a) illustrates the FrankaEmika gripper equipped with sponge based tactile sensors, (b) shows the 3D deformation output of sensors, and (c) represents the mapped force profile of the gripper during grasping and manipulation.}
      \label{seven}
      \vspace{-15pt}
   \end{figure}

In order to achieve an adaptive behaviour and ensure grasp stability for fine manipulation, the sponge based tactile sensors are installed at the jaws of the parallel gripper, as shown in Fig. \ref{seven}(a). These cube-shaped sensors have 16 measuring points and generate displacement data primarily in the z-direction at $115.2$ kHz \cite{c29}. But, to obtain their 3D deformation values to be used for slip detection, a cubic-spline interpolation is applied and the resultant mean output is shown in Fig. \ref{seven}(b). Hence, to map these 3D deformation values to the force profile of the gripper, a shallow neural network (with $5$ hidden neurons) is trained and tested on $100$ and $20$ data samples respectively. This data is collected by tele-operating the gripper to grasp a set of daily life objects (i.e., those mentioned in the confusion matrix of the classifier in Fig. \ref{confusion}) using the grasp stability criteria. The mapped force profile of the Franka-Emika gripper is shown in Fig. \ref{seven}(c). Since, the mapped force values are in sensor frame and need to be transformed into the robot base frame using

\begin{equation}
    {f_c}^{b} = {T_e}^{b}{T_s}^{e}{f_c}^{s}
\label{eqt_12}
\end{equation} 

where ${f_c}^{s}=Neural(D_x, D_y, D_z)$ is the contact force vector (as defined for the force based tasks in the previous subsection), with $D_x$, $D_y$ and $D_z$ representing 3D deformation of the tactile sensors. ${T_e}^{b}\in{SE(3)}$ is the transformation between the end-effector and the robot base frames.

In order to avoid slippage of object (grasped by the gripper) during a human intervention, a modified friction cone criteria is imposed i.e., $\frac{D_z}{\sqrt{{D_x}^{2}+{D_y}^{2}}}>\mu$, where $\mu$ is a co-efficient of friction and its value is tuned empirically based on the nature of task and the system dynamics. For our experimental setup, it is set to $0.75$ (an average value of all the trials) and if this condition is satisfied, the slippage occurs. This tactile information is exploited by the primary forced based tasks to regulate the magnitude of interaction efforts and as well as to switch between the grasping and manipulation phases without violating the defined friction cone criteria.
The overall implementation and execution of the proposed framework is summarized in Algorithm \ref{algorithm}. 

\section{Experimental Evaluation and Discussion}
\begin{algorithm}[t]
\SetAlgoLined
\Input{${P(q), P_i(t), O_c, D_i}$}  
\Output{${\Dot{q}}$}
 \ForEach{${\chi_0}^{d}-{\chi_0}\longrightarrow{e_0}$}{
 $\begin{array}{l}
    \Dot{q_n} = {\argminA_{\dot{q\in{S_s}}}\frac{1}{2} ||{J_n}\dot{q}-\Omega_n{e_n}||^2}+\frac{1}{2}||\Lambda_n||^2
    \\
    s.t
    \\
    l_n\leq{\Lambda_n}{\Dot{q}}\leq{u_n}
  \end{array}$\\
 \If{P(q) == Home $\&$ $P_i(t)$ == wrist pose}{${P(q)}^d$ = ${[X_wY_wZ_w]}^{T}+4\times|Z_w|$ \\
 $\chi_0=J_0(q)\Dot{q}$
 
 }
 
 \If{P(q) == Pre-grasp $\&$ $O_c$ == candid object}{${P(q)}^d$ = ${P_o}^b$ \\
 $\chi_0 = \small{\begin{bmatrix}
 P^{d}(q)-P(q) \\ \Gamma{\psi_o}(q){\zeta_o}^{d}-\Gamma{\psi_o}^{d}{\zeta_o}(q)-\Gamma{S({\zeta_o}^{d})}{\zeta_o}(q)\end{bmatrix}}$
 
 }
 
 \If{P(q) == Grasp $\&$ $D_i$ == sensor deformity}{${f_c}^{s}=Neural(D_x, D_y, D_z)$ \\
 $\chi_f = {R_s}^{b}(q){f_c}^{s}$
 
 }
 
 \If{P(q) == Manipulate $\&$ ${f_c}^{s}$ $\geq{ threshold}$}{${P(q)}^d$ = ${P_w(q)}+{2\times|Z_w|}$ \\
 $\frac{D_z}{\sqrt{{D_x}^{2}+{D_y}^{2}}}>\mu$\\
 $\chi_f = {R_s}^{b}(q){f_c}^{s}$
 
 }
 
 }
\caption{Intuitive Stack-of-Tasks (iSoT)}
\label{algorithm}

\end{algorithm}

To examine the performance and potential of the proposed framework, it has been tested on assembly and disassembly tasks in a laboratory environment (as a representative case study) using a leader(human)--follower(robot) collaborative model. The experimental setup consists of a human subject, a 7-DoF robot arm, a tracking camera, a detection camera, a pair of tactile sensors and the different objects in the scene.

   \begin{figure*}[t]
      \centering
      \includegraphics[width=17cm]{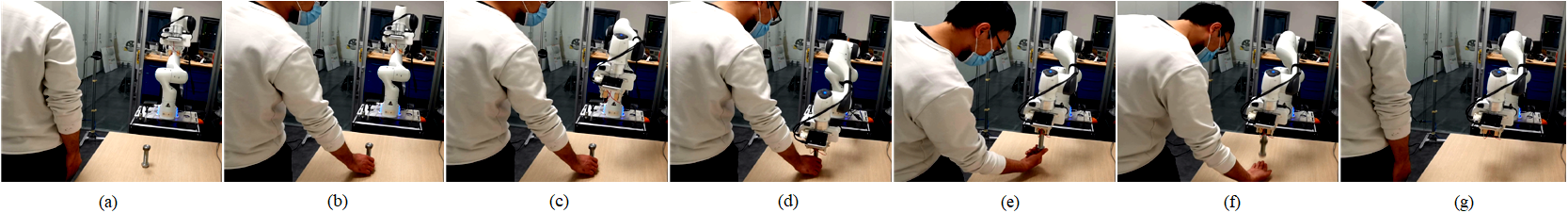}
      \caption{Task 1: Human--robot collaborating on screwing a bolt into a nut: (a) illustrates the homing position of all the actors (object, robot and human subject); in (b) the human subject grasps the bolt from its base; (c) shows a pre-grasping posture of the robot arm; in (d) the robot arm grasps the nut at its center; in (e) the human subject tightens the bolt into the nut while the gripper maintains a stable grasp; in (f) the human subject shows an open palm posture to the detection camera to release the grasped object; in (g) the human subject assumes back the homing position to command the robot arm to return to its initial configuration.\\}
      \label{nine}
      \vspace{-15pt}
   \end{figure*}

   \begin{figure*}[t]
      \centering
      \includegraphics[width=17cm]{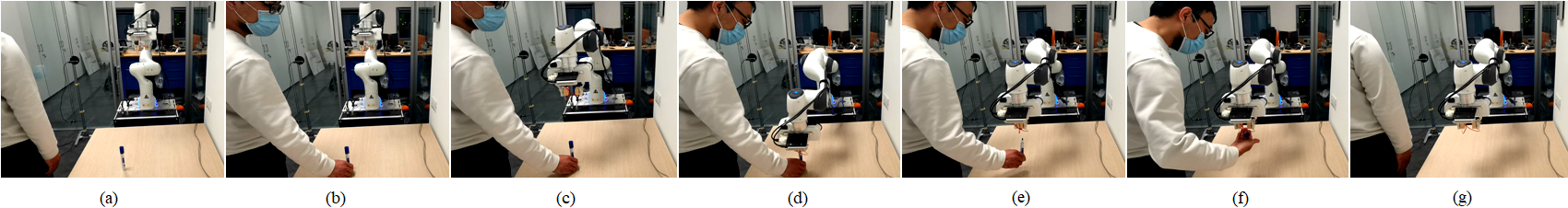}
      \caption{Task 2: Human--robot collaborating on detaching a marker from a cap: (a) illustrates the initial configuration of all the actors in the environment; in (b) the human subject grasps the marker from bottom; (c) shows a pre-grasping posture of the robot arm for object detection; in (d) the robot arm grasps the marker at its cap; in (e) the human subject pulls down the marker while the gripper modulates its force profile to hold its cap stably; in (f) the gripper releases the cap on detecting an open palm posture of the human subject; in (g) the robot arm return to its homing configuration thereby following the posture of human partner.}
      \label{ten}
      \vspace{-15pt}
   \end{figure*}

The tasks shown in Figs. \ref{nine} and \ref{ten} involve screwing a bolt into a nut (assembly) and detaching a marker from its cap (disassembly), respectively. At first, both the agents i.e., the human and robot and also the environment, are in their initial configurations. In these experiments, the human subject (leader) uses his/her right arm to hold the bolt and the marker from underneath, as shown in Figs. \ref{nine}(a) and \ref{ten}(a). This posture of active human arm is monitored and estimated by the installed tracking camera, which commands the robot arm (follower) to move closer to and vertically above the human wrist (to a position four times the height of the human wrist-- a safe detection distance) such that the candidate object gets into the field of view of the eye-in-hand (detection) camera in a pre-grasping configuration, as shown in Figs. \ref{nine}(b) and \ref{ten}(b). Under this view, the detection camera, using the RANSAC algorithm together with the trained SVM classifier, helps to segment-out, localize and determine the pose of the candidate objects (nut-bolt and marker-cap) in the scene. Using this pose information, the robot arm exploits grasping action from the iSoT to grip each of them at their centroids (i.e, at the boundaries of object's surface on an orthogonal line passing through centroid), as shown in Figs. \ref{nine}(c) and \ref{ten}(c). Once the contact is established and detected by the tactile sensors, the manipulation action is called on to lift the grasped object vertically upwards (in the z-direction to a position two times the height of the human wrist-- a minimum clearance distance). This ensures a clearance between the table and object and also provides space for the human subject to undertake further actions onto the grasped object (by the gripper), as shown in Figs. \ref{nine}(d) and \ref{ten}(d). While the human subject is tightening the bolt into the nut and detaching the marker from its cap, as shown in Figs. \ref{nine}(e) and \ref{ten}(e) respectively, the corresponding deformation into the tactile sensors helps to modulate the force profile of the gripper, with reference to (\ref{eqt_12}). Once the human coworker finishes the task, he/she shows an open palm posture to the detection camera as a gesture to command the gripper to release the grasped object, as shown in Figs. \ref{nine}(f) and \ref{ten}(f). Finally, for the next trial or new task, the robot arm returns to its homing configuration thereby following the posture of the active human arm using the tracking camera, as shown in Figs. \ref{nine}(g) and \ref{ten}(g).

   \begin{figure}[t]
      \centering
      \includegraphics[width=8.5cm]{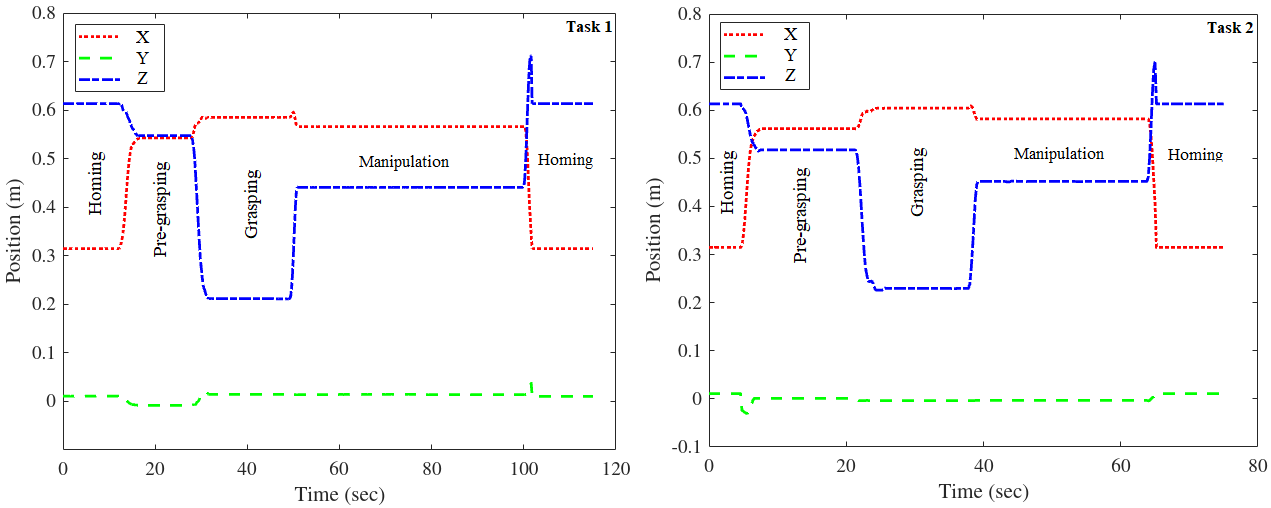}
      \caption{End-effector trajectories of the robot arm at different levels of the iSoT during both the co-manipulation scenarios.}
      \label{eleven}
      \vspace{-15pt}
   \end{figure}
   
\begin{figure}[t]
      \centering
      \includegraphics[width=8.5cm]{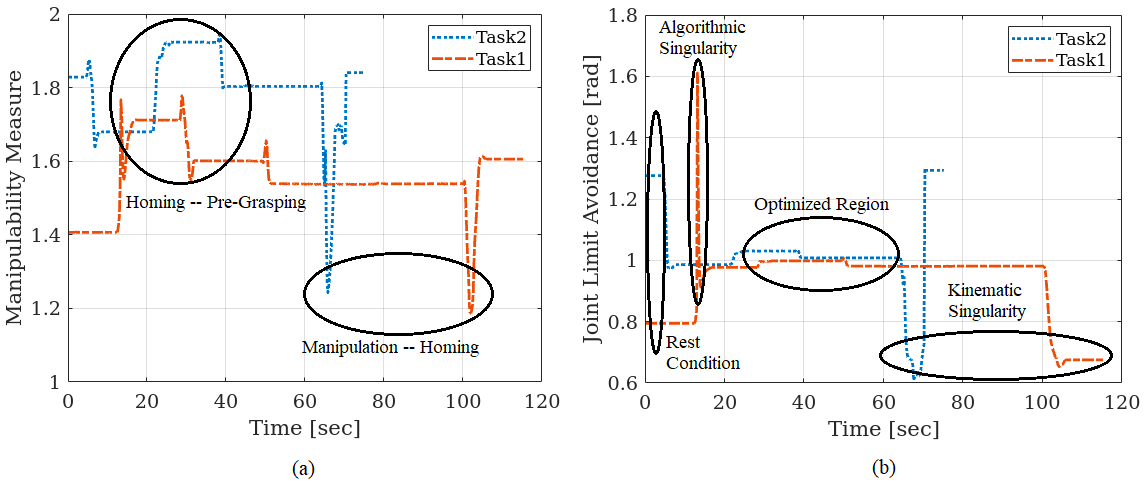}
      \caption{Quantitative evaluation of secondary (performance) tasks for both the co-manipulation scenarios, (a) shows net values of manipulability index as a function of robot arm configurations at different instants, and (b) illustrates overall deviations into the robot arm joints from their mid-point values during task execution.}
      \label{elevens}
      \vspace{-15pt}
   \end{figure}

The complete state-action trajectory of the robot arm (follower) for both the iHRC tasks at different instants is shown in Fig. \ref{eleven}. It can be seen that the nature and duration of both the tasks is however different (i.e., the assembly task takes longer time than the disassembly task). But, the proposed framework is promisingly robust against the task variations and also ensures consistent actions (i.e., repetition of desired movements) of the robot arm under same (during trials) as well as under different (for distinct tasks) conditions. Moreover, the quantitative values of performance indices that have been considered as secondary tasks (i.e, manipulability measure and joint limit avoidance) to deal with the robot arm redundancy (i.e, $n=7 > m=5$), are shown in Fig. \ref{elevens}. It can be seen from Fig. \ref{elevens}(a) that during both the co-manipulations scenarios, the iSoT framework while switching from homing to pre-grasping configurations, tries to maximize the manipualbility measure (i.e, a jump of 0.1), and regulates it to the nominal values (i.e, 1.6 and 1.8) during a transition from manipulation to homing configurations, where the robot arm passes near to the kinematic singularity. Similarly, the solutions obtained from iSoT framework also try to maintain the robot arm configurations close to the mid-point of their mechanical joint limits (i.e, optimized region) with minor overshoots (i.e, within the bounds defined in \cite{c30}) due to the algorithmic and kinematic singularities during the intuitive switching, as shown in Fig. \ref{elevens}(b).           

\begin{table*}[t]
\caption{Comparative analysis of collaborative scenarios using different evaluation metrics.}
\begin{tabular}{|P{11em}|P{11.2em}|P{11.2em}|P{11.2em}|P{11.2em}|}
    \hline
\multirow{2.3}{*}{\textbf{Evaluation Metric}}
    & \multicolumn{2}{c|}{\textbf{Human--Human Cooperation}}
        & \multicolumn{2}{c|}{\textbf{Human--Robot Collaboration}}     \\
    \cline{2-5}
    & \textbf{Task 1}
        & \textbf{Task 2}
            & \textbf{Task 1}   & \textbf{Task 2}                                   \\
    \hline
\makecell{Approach Adaptation \\(m $\&$ rad)}
    & $\Delta$r: ($\mu$=0.5961, $\sigma$=0.0052) $\Delta$$\theta$: ($\mu$=0.1414, $\sigma$=0.0025)    & $\Delta$r: ($\mu$=0.6388, $\sigma$=0.0034) $\Delta$$\theta$: ($\mu$=0.1992, $\sigma$=0.0036)    & $\Delta$r: ($\mu$=0.6112, $\sigma$=0.0021) $\Delta$$\theta$: ($\mu$=0.1396, $\sigma$=0.0014) & $\Delta$r: ($\mu$=0.6525, $\sigma$=0.0071) $\Delta$$\theta$: ($\mu$=0.1921, $\sigma$=0.0033)        \\
    \hline
\makecell{Task Coordination \\Latency (sec)}
     & 3.8    & 2.7    & 27.8 & 23        \\
    \hline
\makecell{Grasp Correction \\(mm)}
     & 18.9048    & 16.1376    & 1.955 & 1.181     \\
    \hline
\makecell{Cumulative Posture \\Deviation ($\%$)}
     & 5.0929    & 3.8655    & 0.1169 & 0.0993        \\
    \hline
\makecell{Task Repeatability \\(C)}
     & 0.6102   & 0.6612    & 0.9142 & 0.9607        \\
    \hline
    \end{tabular}
    \\ \\
\vspace{-15pt}
\label{quantitative}
\end{table*}

It must be noted that the existing settings for pre-grasping and manipulation configurations (i.e., four and two times the height of the human wrist respectively) in the proposed framework are chosen empirically by taking into account the robot arm singularities, the human arm stability, the nature of objects, and the strength of gripper’s force. However, this constraint can be relaxed by considering two potential options: (1) re-formulating the proposed framework in either hybrid or shared architecture, instead of traded, to exploit all the sensory feedback simultaneously for instant decision-making. This may come at the cost of higher computational complexity and communication latency, and as well as requires to harmonize the perception (i.e, $5$ Hz, $25$ Hz, 115.2 kHz) and action (i.e, $1$ kHz) loop rates, and (2) incorporating a force control (either impedance or admittance) profile for the robot arm to provide flexibility and ease to a human coworker to adjust its end-effector’s position according to their requirements and the task dimensions. Unfortunately, this may disturb the intuitive leader-follower paradigm in the HRC scenarios and also poses additional constraints on the tuning of its parameters i.e, stiffness/compliance ($K_m/C_m$) and damping ($D_m$). In the future, the proposed framework shall be upgraded to make a trade-off between the controller complexity and interaction flexibility for a run-time decision-making to further optimize the human comfort and the task ergonomics.

\section{Preliminary Comparative Case Study}

The proposed framework is designed for intuitive human--robot collaboration, where a robot could assist a human in doing routine assembly/disassembly tasks. It is therefore important to validate the functionality and usability of the framework against the traditional human--human cooperation for such tasks that are prevalent in the industry nowadays. 

A preliminary comparative experiment was designed to analyze the task performance with and without the proposed framework. The experiment was structured to have the same leader--follower setting, with a human always in the leader role, and the follower role being interchanged between a human and the robot. Figs. \ref{nine}, \ref{ten} and \ref{twelve} show the setup of the experimental evaluation. In the setup, to maintain consistency of the tasks, the leader and follower (during human-human and human-robot scenarios) were kept $r=0.88$ m, $\theta=0.6457$ rad apart (i.e., the polar distance between the origin of their reference frames, as shown in Fig. \ref{four}) besides the wooden table. The two tasks considered are: (Task 1) screwing a bolt into a nut, and (Task 2) detaching a marker from a cap, as shown in Figs. \ref{nine} and \ref{ten}. Both the tasks were repeated fives times each. During the trials, the objects were placed in slightly different poses but respecting the kinematic reach of the robot arm (i.e., $r=0.85$ m, $\theta=4.7124$ rad). This helps not only to understand how robust the proposed framework is, but also how well the leader and follower adapt to their coordinated movements against the environmental variations. The task instructions for this comparative experiment were as follows:

\begin{itemize}
    \item Both the leader and follower start in homing configuration during each trial for both the tasks.
    \item Leader, using their dominant arm, grips the candidate object at its base.
    \item Follower perceives the environment, as well as the leader’s actions, and moves its arm accordingly over the object in a pre-grasping posture. 
    \begin{itemize}
        \item For the human follower, the arm is the non-dominant hand. This allows a one-to-one comparison with the robot arm, which acts as the non-dominant hand in the human--robot co-manipulation scenario.
        \item Follower brings their arm over the object at a height of four times the position of the leader wrist. 
    \end{itemize}
    \item Follower grasps the object from its head provided that the leader is already gripping it from underneath.
    \item Follower lifts the object up to provide space to leader for further interactions with it.
    \begin{itemize}
        \item The object is lifted vertically (in z-direction) to two times the height of the leader wrist position. 
    \end{itemize}
    \item Leader co-manipulates the object according to the task requirements.
    \item Leader shows an open palm posture to follower as a gesture to release the grasped object. 
    \item Leader returns to homing position, followed by the follower for the next trial.
\end{itemize}

For the comparative case study, five distinct evaluation metrics are considered, as reported in Table \ref{quantitative} and are elaborated as follows:

   \begin{figure*}[t]
      \centering
      \includegraphics[width=17cm]{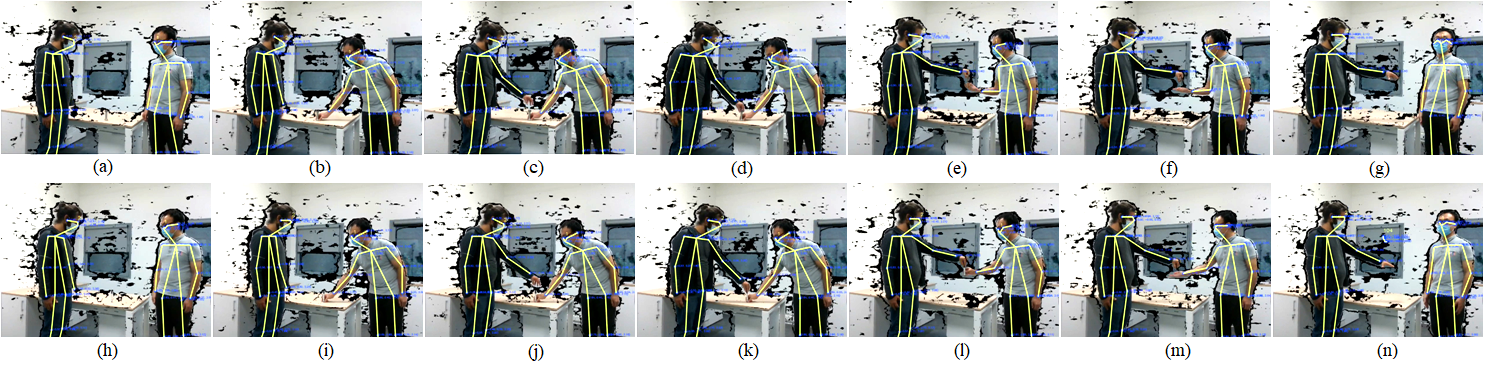}
      \caption{Postures of human-human partners collaborating on performing the assembly and disassembly tasks in the laboratory environment (using the skeleton tracking algorithm): (a--g) represent the human partners for screwing bolt into nut task; (h--n) show the human subjects performing the dismounting marker from cap task. }
      \label{twelve}
      \vspace{-15pt}
   \end{figure*}
   
\begin{enumerate}
\item Approach Adaptation --- quantifies the change in follower's wrist position in approaching the candidate object from homing to pre-grasping configurations. This metric shows a mean value ($\mu$) and standard deviation ($\sigma$) of polar distance (r, $\theta$), measured in (meters, radians). The distance is calculated as a difference in positions of follower's wrist at homing and pre-grasping configurations.
\item Task Coordination Latency --- measures the total time lag of follower's response to leader's actions and sensory feedback for the task execution through all the phases -- homing, pre-grasping, grasping, and manipulation. This metric shows an average value of time delay over the trials, measured in (seconds), by the follower in responding to the leader's requests and change of environment, that are detected using the perceptual feedback.
\item Grasp Correction --- defines the adjustment in precision grasp established by follower for stabilization of the candidate object from grasping to manipulation phases over the trials of task. This metric shows a change in contacts' position of follower's grasp, measured in (millimeters), to avoid objects' slippage during co-manipulation. 
\item Cumulative Posture Deviation --- characterizes a relative variation into overall trajectory of follower's wrist from homing to manipulation during task execution. This metric shows a ratio of change in follower's wrist position, measured in (percentage), during the consecutive trials.
\item Task Repeatability --- specifies a similarity in follower's postures in repeating the same task over the trials from grasping to manipulation. This metric shows consistency of follower's actions from homing to manipulation configurations, measured using the consistency degree (C $\in$ \{0,1\}), for the same tasks over the five repetitions. Higher value of C defines a greater similarity over the trials.  
\end{enumerate}

The values of evaluation metrics given in Table \ref{quantitative} are computed from the data that is collected using the same skeleton tracking algorithm discussed in previous subsection for gesture monitoring and posture evaluation. It is evident from Table \ref{quantitative} that the two considered tasks, when performed using the proposed framework give a comparable performance in approaching the candidate objects as to human-human cooperation (i.e., with a diminutive difference in their standard deviations of less than $0.004$ and $0.002$ in the values of $r$ and $\theta$ respectively). This behaviour is achieved due to the intuitive formulation of the robot's actions (iSoT) using multi-sensory information that help to effectively perceive the collaborative environment. Moreover, the proposed framework, on replacing the human follower with the robot arm, results in a better task execution and repeatability (i.e., with $0.9142$ C and $0.9607$ C), as wells an improved precision in object grasping (i.e., a small contact correction of $1.955$ mm and $1.181$ mm) and least deviations into the follower's actions (i.e., $0.1169\%$ and $0.0993\%$) during the trials of both the tasks. These characteristics are highly desirable in industry for improving the quality of end products \cite{c31}. However, the proposed framework shows an increased time lag (i.e., $27.8$ sec and $23$ sec) in the overall task execution due to a communication latency between the action (robot arm) and perception (sensory feedback) loops. Also, it is noticeable from Table \ref{quantitative} that the Task 2, due to its simplicity and minimalism, has more promising results for all the evaluation metrics than the Task 1, under both the collaborative scenarios.   

Hence, it can be concluded that the proposed framework is robust and efficient in executing and repeating the tasks with least deviations in overall follower's actions and greater precision in object grasping (i.e., contact formation). Also, the proposed framework ensures desired accuracy in approaching the candidate objects, against their varying poses during the trials, similar to human substitute (follower). But unfortunately, it suffers from coordination delay due to accumulative latency between the sensory feedback and the control loop of the robot arm. However, the advances into the fusing strategies, learning algorithms and processing capabilities for the multi-sensing modalities and instant decision-making are starting to resolve some of these issues to a certain extent \cite{c32}\cite{c33}\cite{c34}.      

In this preliminary study, two human subjects participated, one as leader and other as follower. Both were PhD students, who have worked with collaborative robots but only in laboratory environment. In the future, the analysis will be extended to professional and experienced individuals to decipher the industrial and commercial applications and possible implications of the proposed framework.

\section{CONCLUSIONS}

A control framework, for the collaborative fine manipulation tasks, that follows the idea of stack-of-tasks (SoT) using constrained quadratic programming (QP) is proposed in this research. The framework is highly intuitive because of the natural insights that can be derived from the human-arm gestures and task progression using the depth and haptic feedback in a traded fashion. The depth information from the tracking and detection cameras, estimate the human-arm postures and objects' poses, using the skeleton tracking and semantic segmentation algorithms respectively. The haptic information from the tactile sensors is used to modulate the force profile of the gripper to achieve an adaptive interaction during co-manipulation of objects by considering the friction cone criteria. The performance of the proposed framework is examined while performing the assembly and disassembly HRC tasks at laboratory environment using a human(leader)-robot(follower) partnership. The robustness and optimality of the proposed framework are assessed against the human-human cooperation for the similar tasks using distinct evaluation metrics i.e., approach adaptation, grasp correction, task coordination latency, cumulative posture deviation, and task repeatability. The results obtained confirm the potential of the proposed framework in executing and repeating the collaborative tasks with more accuracy and precision while suffers from communication latency. The accompanied media file shows the demonstrations of the considered tasks under both the collaborative paradigms.  

The proposed framework will be applied to field agricultural robotics for precision manipulation i.e., extending the research discussed in \cite{c35}. In the future, the framework will be upgraded to include the incremental learning that will help in switching the role between the human and robot (i.e., either can be leader or follower) based on the nature of task and the degree of collaboration needed.

\bibliographystyle{IEEEtran}

\begin{thebibliography}{99}

\bibitem{c1} O. Khatib, K. Yokoi, O. Brock, K. Chang, and A. Casal, “Robots in human environments: Basic autonomous capabilities,” The International Journal of Robotics Research, vol. 18, no. 7, pp. 684–696, 1999.

\bibitem{c2} A. De Santis, B. Siciliano, A. De Luca, and A. Bicchi, “An atlas of physical human–robot interaction,” Mechanism and Machine Theory, vol. 43, no. 3, pp. 253–270, 2008.

\bibitem{c3} M. Raessa, J. C. Y. Chen, W. Wan, and K. Harada, “Human-in-the-loop robotic manipulation planning for collaborative assembly,” IEEE Transactions on Automation Science and Engineering, vol. 17, no. 4, pp. 1800–1813, 2020.

\bibitem{c4} I. Maurtua, A. Ibarguren, J. Kildal, L. Susperregi, and B. Sierra, “Human–robot collaboration in industrial applications: Safety, interaction and trust,” International Journal of Advanced Robotic Systems, vol. 14, no. 4, p. 1729881417716010, 2017.

\bibitem{c5} A. De Santis, V. Lippiello, B. Siciliano, and L. Villani, “Human-robot interaction control using force and vision,” in Advances in Control Theory and Applications. Springer, 2007, pp. 51–70.

\bibitem{c6} X. Yu, W. He, Q. Li, Y. Li, and B. Li, “Human-robot co-carrying using visual and force sensing,” IEEE Transactions on Industrial Electronics, vol. 68, no. 9, pp. 8657–8666, 2020.

\bibitem{c7} Y. Zhang, G. Zhang, Y. Du, and M. Y. Wang, “Vtacarm. a vision-based tactile sensing augmented robotic arm with application to human-robot interaction,” in 2020 IEEE 16th International Conference on Automation Science and Engineering (CASE). IEEE, 2020, pp. 35–42.

\bibitem{c8} M. Khatib, K. Al Khudir, and A. De Luca, “Visual coordination task for human-robot collaboration,” in 2017 IEEE/RSJ international conference on intelligent robots and systems (IROS). IEEE, 2017, pp. 3762–3768.

\bibitem{c9} H. Liu, T. Fang, T. Zhou, and L. Wang, “Towards robust human-robot collaborative manufacturing: Multimodal fusion,” IEEE Access, vol. 6, pp. 74 762–74 771, 2018.

\bibitem{c10} P. A. Schmidt, E. Maël, and R. P. Würtz, “A sensor for dynamic tactile information with applications in human–robot interaction and object exploration,” Robotics and Autonomous Systems, vol. 54, no. 12, pp. 1005–1014, 2006.

\bibitem{c11} S. Scheggi, F. Morbidi, and D. Prattichizzo, “Human-robot formation control via visual and vibrotactile haptic feedback,” IEEE Transactions on Haptics, vol. 7, no. 4, pp. 499–511, 2014.

\bibitem{c12} A. Cherubini and D. Navarro-Alarcon, “Sensor-based control for collaborative robots: Fundamentals, challenges, and opportunities,” Frontiers in Neurorobotics, vol. 14, p. 113, 2021.

\bibitem{c13} Y. Maeda, T. Hara, and T. Arai, “Human-robot cooperative manipulation with motion estimation,” in Proceedings 2001 IEEE/RSJ International Conference on Intelligent Robots and Systems. Expanding the Societal Role of Robotics in the the Next Millennium (Cat. No. 01CH37180), vol. 4. Ieee, 2001, pp. 2240–2245.

\bibitem{c14} A. Cherubini, R. Passama, A. Meline, A. Crosnier, and P. Fraisse, “Multimodal control for human-robot cooperation,” in 2013 IEEE/RSJ International Conference on Intelligent Robots and Systems. IEEE, 2013, pp. 2202–2207.

\bibitem{c15} D. J. Agravante, A. Cherubini, A. Bussy, P. Gergondet, and A. Kheddar, “Collaborative human-humanoid carrying using vision and haptic sensing,” in 2014 IEEE international conference on robotics and automation (ICRA). IEEE, 2014, pp. 607–612.

\bibitem{c16} A. Cherubini, R. Passama, P. Fraisse, and A. Crosnier, “A unified multimodal control framework for human–robot interaction,” Robotics and Autonomous Systems, vol. 70, pp. 106–115, 2015.

\bibitem{c17} E. Magrini, F. Ferraguti, A. J. Ronga, F. Pini, A. De Luca, and F. Leali, “Human-robot coexistence and interaction in open industrial cells,” Robotics and Computer-Integrated Manufacturing, vol. 61, p. 101846, 2020.

\bibitem{c18} M. Costanzo, G. De Maria, and C. Natale, “Handover control for human-robot and robot-robot collaboration,” Frontiers in Robotics and AI, vol. 8, p. 132, 2021.

\bibitem{c19} S. Grushko, A. Vysocky, P. Oscadal, M. Vocetka, P. Novak, and Z. Bobovsky, “Improved mutual understanding for human-robot collaboration: Combining human-aware motion planning with haptic feedback devices for communicating planned trajectory,” Sensors, vol. 21, no. 11, p. 3673, 2021.

\bibitem{c20} B. Siciliano, L. Sciavicco, L. Villani, and G. Oriolo, Modelling, planning and control. Springer, 2009.

\bibitem{c21} J. Cacace, F. Ruggiero, and V. Lippiello, “Hierarchical task-priority control for human-robot co-manipulation.” in HFR, 2019, pp. 125–138.

\bibitem{c22} W. Kim, J. Lee, L. Peternel, N. Tsagarakis, and A. Ajoudani, “Anticipatory robot assistance for the prevention of human static joint overloading in human–robot collaboration,” IEEE robotics and automation letters, vol. 3, no. 1, pp. 68–75, 2017.

\bibitem{c23} C. Gaz, E. Magrini, and A. De Luca, “A model-based residual approach for human-robot collaboration during manual polishing operations,” Mechatronics, vol. 55, pp. 234–247, 2018.

\bibitem{c24} F. Caccavale, V. Lippiello, G. Muscio, F. Pierri, F. Ruggiero, and L. Villani, “Grasp planning and parallel control of a redundant dual-arm/hand manipulation system,” Robotica, vol. 31, no. 7, pp. 1169–1194, 2013.

\bibitem{c25} A. Rocchi, E. M. Hoffman, E. Farnioli, and N. G. Tsagarakis, “A whole-body stack-of-tasks compliant control for the humanoid robot coman,” in IEEE/RSJ international conference on intelligent robots and systems (IROS 2015), Hamburg, Germany, vol. 28, 2015.

\bibitem{c26} Cubemos, “Mehlbeerenstrasse 2, 82024 Taufkirchen (Munich) Germany”, \url{https://www.cubemos.com/skeleton-tracking-sdk}, 2021, [Online;accessed 19-July-2021].

\bibitem{c27} F. Flacco, T. Kroeger, A. De Luca, and O. Khatib, “A depth space approach for evaluating distance to objects,” Journal of Intelligent and Robotic Systems, vol. 80, no. 1, pp. 7–22, 2015.

\bibitem{c28} S. Katyara, F. Ficuciello, F. Chen, B. Siciliano, and D. G. Caldwell, “Vision based adaptation to kernelized synergies for human inspired robotic manipulation,” arXiv preprint arXiv:2012.07046, 2020.

\bibitem{c29} Touchence, “2-21-10 Kitaueno, Taito-ku, Tokyo 110-0014,” \url{http://www.touchence.jp/en/products/cube.html}, 2021, [Online; accessed 19-July-2021].

\bibitem{c30} “Franka Emika robot and interface specifications,” \url{https://frankaemika.
github.io/docs/control parameters.html}, accessed: 2022-01-04.

\bibitem{c31} A. Ajoudani, A. M. Zanchettin, S. Ivaldi, A. Albu-Schäffer, K. Kosuge, and O. Khatib, “Progress and prospects of the human–robot collaboration,” Autonomous Robots, vol. 42, no. 5, pp. 957–975, 2018.

\bibitem{c32} J. Ilonen, J. Bohg, and V. Kyrki, “Fusing visual and tactile sensing for 3-d object reconstruction while grasping,” in 2013 IEEE International Conference on Robotics and Automation. IEEE, 2013, pp. 3547–3554.

\bibitem{c33} O. B. Kroemer, R. Detry, J. Piater, and J. Peters, “Combining active learning and reactive control for robot grasping,” Robotics and Autonomous systems, vol. 58, no. 9, pp. 1105–1116, 2010.

\bibitem{c34} M. Peniak, “Gpu computing for cognitive robotics,” 2014.

\bibitem{c35} S. Katyara, F. Ficuciello, D. G. Caldwell, F. Chen, and B. Siciliano, “Reproducible pruning system on dynamic natural plants for field agricultural robots,” in International Workshop on Human-Friendly Robotics. Springer, 2020, pp. 1–15.

\end{thebibliography}

\end{document}